\documentclass[10pt,twocolumn,letterpaper]{article}

\usepackage{cvpr}
\usepackage{times}
\usepackage{epsfig}
\usepackage{graphicx}
\usepackage{amsmath}
\usepackage{amssymb}
\usepackage{subcaption}
\usepackage{multirow}

\usepackage[breaklinks=true,bookmarks=false,colorlinks]{hyperref}
\usepackage{cleveref}

\cvprfinalcopy 

\def\cvprPaperID{9} 

\begin{document}

\title{Temporal Attentive Alignment for Video Domain Adaptation}

\author{Min-Hung Chen \hspace{3em}
Zsolt Kira \hspace{3em}
Ghassan AlRegib \\
Georgia Institute of Technology\\
{\tt\small \{cmhungsteve, zkira, alregib\}@gatech.edu}
}

\maketitle

\begin{abstract}
   Although various image-based domain adaptation (DA) techniques have been proposed in recent years, domain shift in videos is still not well-explored. Most previous works only evaluate performance on small-scale datasets which are saturated. Therefore, we first propose a larger-scale dataset with larger domain discrepancy: \textbf{UCF-HMDB$\boldsymbol{_{full}}$}. Second, we investigate different DA integration methods for videos, and show that simultaneously aligning and learning temporal dynamics achieves effective alignment even without sophisticated DA methods. Finally, we propose \textbf{Temporal Attentive Adversarial Adaptation Network (TA$\boldsymbol{^3}$N)}, which explicitly attends to the temporal dynamics using domain discrepancy for more effective domain alignment, achieving state-of-the-art performance on three video DA datasets. The code and data are released at \url{http://github.com/cmhungsteve/TA3N}.
\end{abstract}

\section{Introduction}

\begin{figure}[!t]
\centering
\includegraphics[width=0.475\textwidth]{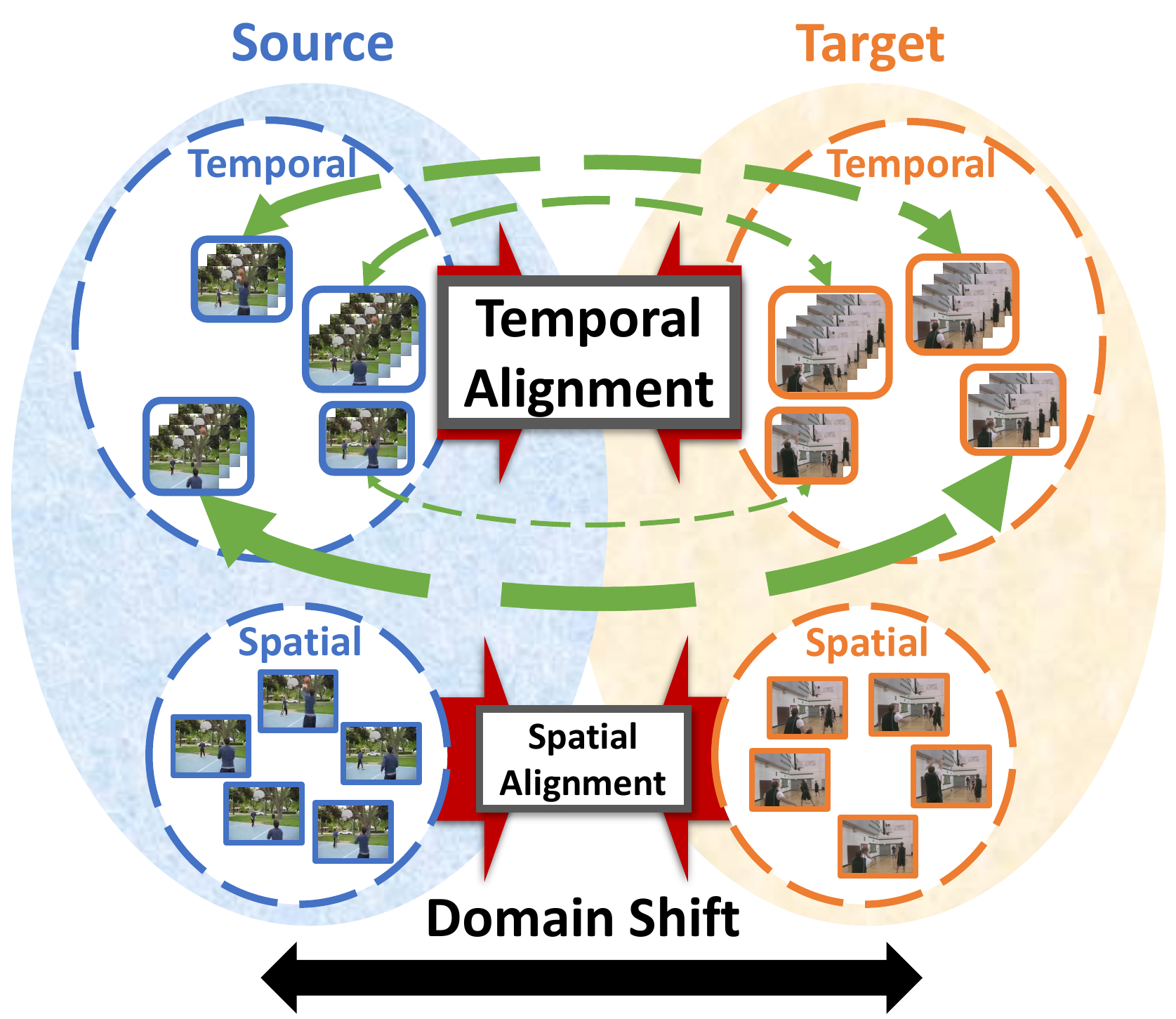}
\caption{An overview of proposed TA$^3$N for video DA. 
In addition to spatial discrepancy between frame images, videos also suffer from temporal discrepancy between sets of time-ordered frames that contain multiple local temporal dynamics with different contributions to the overall domain shift, as indicated by the thickness of green dashed arrows. 
Therefore, we propose to focus on aligning the temporal dynamics which have higher domain discrepancy using a learned attention mechanism to effectively align the temporal-embedded feature spaces for videos.
Here we use the action \textit{basketball} as the example.}
\label{fig:overview_video_DA}
\end{figure}

Unsupervised \textit{Domain adaptation (DA)}~\cite{csurka2017comprehensive} has been studied extensively in recent years to address the \textit{domain shift} problem~\cite{quionero2009dataset}, which means the models trained on source labeled datasets do not generalize well to target datasets and tasks, without access to any target labels.
While many DA approaches are able to diminish the distribution gap between source and target domains while learning discriminative deep features~\cite{ganin2015unsupervised, long2017deep, li2018adaptive, saito2018maximum}, most methods have been developed only for images and not videos.

Furthermore, unlike image-based DA work, 
there do not exist well-organized datasets to evaluate and benchmark the performance of DA algorithms for videos. 
The most common datasets are \textit{UCF-Olympic} and \textit{UCF-HMDB$_{small}$}~\cite{sultani2014human, xu2016dual, jamal2018deep}, which have only a few overlapping categories between source and target domains. 
This introduces limited domain discrepancy so that a deep CNN architecture can achieve nearly perfect performance even without any DA method (details in \Cref{sec:experimental_results} and \Cref{table:sota_ucf-olympic_ucf-hmdb-small}).
Therefore, we propose a larger-scale video DA dataset, \textit{UCF-HMDB$_{full}$}, by collecting all relevant and overlapping categories between UCF101~\cite{soomro2012ucf101} and HMDB51~\cite{kuehne2011hmdb}, leading to three times larger than previous two datasets, and contains larger domain discrepancy (details in \Cref{sec:experimental_results} and \Cref{table:sota_ucf-hmdb_full}). 

Videos can suffer from domain discrepancy along both the spatial and temporal directions, bringing the need of alignment for embedded feature spaces along both directions, as shown in \Cref{fig:overview_video_DA}. 
However, most DA approaches have not explicitly addressed the domain shift problem in the temporal direction.
Therefore, we first investigate different DA integration methods for video classification and 
propose \textit{Temporal Adversarial Adaptation Network (TA$^2$N)}, which simultaneously aligns and learns temporal dynamics, to effectively align domains spatio-temporally, outperforming other approaches which naively apply more sophisticated image-based DA methods for videos. 
To further achieve more effective temporal alignment, we propose to focus more on aligning those which have higher contribution to the overall domain shift, such as the local temporal features connected by thicker green arrows shown in \Cref{fig:overview_video_DA}.
Therefore, we propose \textbf{Temporal Attentive Adversarial Adaptation Network (TA$\boldsymbol{^3}$N)} to explicitly attend to the temporal dynamics by taking into account the domain distribution discrepancy,
achieving state-of-the-art performance on all three investigated video DA datasets. (\href{http://github.com/cmhungsteve/TA3N}{GitHub})

\section{Technical Approach}
Here we introduce our baseline model (\Cref{sec:baseline}) and proposed methods for video DA (\Cref{sec:TAAN,sec:TAAAN}).

\subsection{Baseline Model} \label{sec:baseline}
\begin{figure}[!ht]
\centering
\includegraphics[scale=0.373]{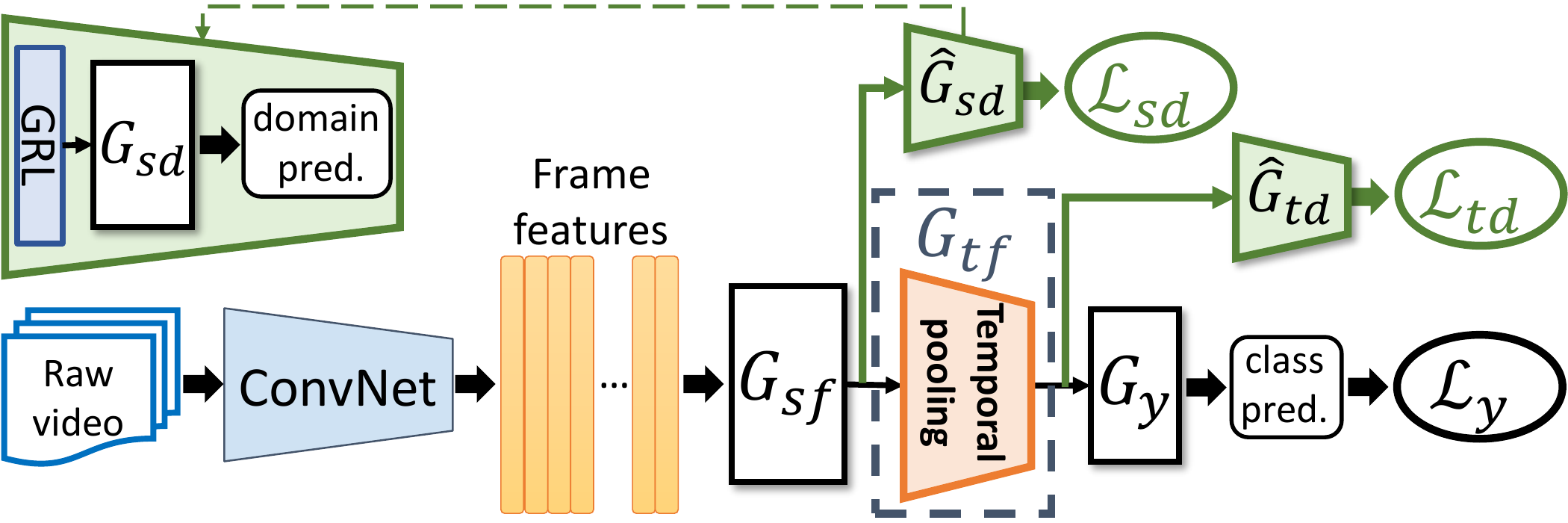}
\caption{Baseline architecture (TemPooling) with the adversarial discriminators $\hat{G}_{sd}$ and $\hat{G}_{td}$. $\mathcal{L}_y$ is the class prediction loss, and $\mathcal{L}_{sd}$ and $\mathcal{L}_{td}$ are the domain losses. 
}
\label{fig:TemPooling_RevGrad}
\end{figure}

Given the recent success of large-scale video classification using CNNs~\cite{karpathy2014large}, we build our baseline on such architectures, as shown in the lower part of Figure~\ref{fig:TemPooling_RevGrad}. 

We first feed the 
frame-level feature representations extracted from ResNet~\cite{he2016deep} pre-trained on ImageNet into our model, 
which can be divided into two parts: 1) \textit{Spatial module $G_{sf}$}, 
which consists of multilayer perceptrons (MLP) that aims to convert the general-purpose feature vectors into task-driven feature vectors, where the task is video classification in this paper; 
2) \textit{Temporal module $G_{tf}$}, which aggregates the frame-level feature representations to form a single video-level feature representation for each video. Our baseline architecture conducts mean-pooling along the temporal direction in $G_{tf}$, so we note it as \textit{TemPooling}.

To address domain shift, we integrate TemPooling with \textit{Adversarial Discriminator $\hat{G}_d$}, which is the combination of a gradient reversal layer (GRL) and a domain classifier $G_d$, inspired by \cite{ganin2015unsupervised}. Through adversarial training, $G_d$ is optimized to discriminate data across domains, while the feature generator is optimized to gradually align the feature distributions across domains. 
$\hat{G}_d$ is integrated in two ways: 1) $\hat{G}_{sd}$: show how directly applying image-based DA approaches can benefit video DA; 2) $\hat{G}_{td}$: indicate how DA on temporal-dynamics-encoded features benefits video DA. 

\subsection{Integration of Temporal Dynamics with DA} \label{sec:TAAN}
One main drawback of directly integrating image-based DA approaches into TemPooling
is that the relation between frames is missing. 
Therefore, we would like to address this question: \textit{Does the video DA problem benefit from encoding temporal dynamics into features?} 

Given the fact that humans can recognize actions by reasoning the observations across time, we propose the \textit{TemRelation} architecture by replacing the temporal pooling mechanism in $G_{tf}$ with the Temporal Relation module, which is modified from \cite{zhou2018temporalrelation}, as shown in Figure~\ref{fig:TemRelation_RevGrad_TransAttn}. 
Specifically, We fuse the time-ordered feature representations with MLPs into $n$-frame relation $R_n$, and then sum up all $R_n$ into the final video representation to capture temporal relations at multiple time scales. 
To align and encode temporal dynamics simultaneously instead of solely modifying the Temporal module, we propose \textbf{Temporal Adversarial Adaptation Network (TA$^2$N)},
which integrates relation discriminators $\hat{G}^n_{rd}$ \textit{inside} the Temporal module with corresponding $n$-frame relations to properly align different temporal relations,  
outperforming those which are extended from sophisticated image-based DA approaches although TA$^2$N is adopted from a simpler DA method (details in \Cref{table:sota_ucf-hmdb_full}).

\subsection{Temporal Attentive Alignment for Videos} \label{sec:TAAAN}
\begin{figure}[!t]
\centering
\includegraphics[width=0.475\textwidth]{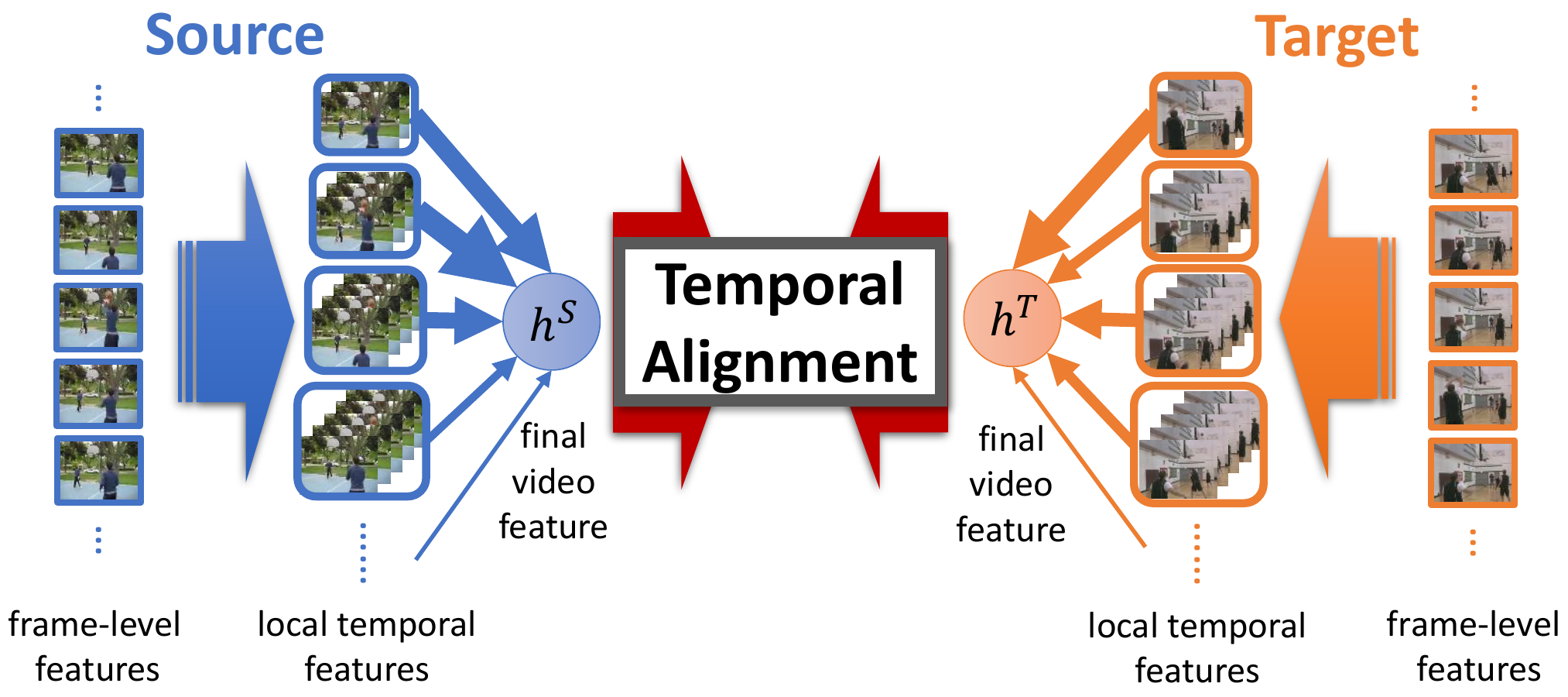}
\caption{The domain attention mechanism in TA$^3$N.
Thicker arrows corresponds to larger attention weights.
}
\label{fig:overview_domain_attention}
\end{figure}

Although aligning temporal features across domains benefits video DA, not all the features equally contribute to the overall domain shift. 
Therefore, we propose to focus more on aligning those which have larger domain discrepancy and assign them with 
larger attention weighting values, as shown in \Cref{fig:overview_domain_attention}.
The main question becomes: \textit{How to incorporate domain discrepancy for attention?} 

\begin{figure}[!t]
\centering
\includegraphics[width=0.475\textwidth]{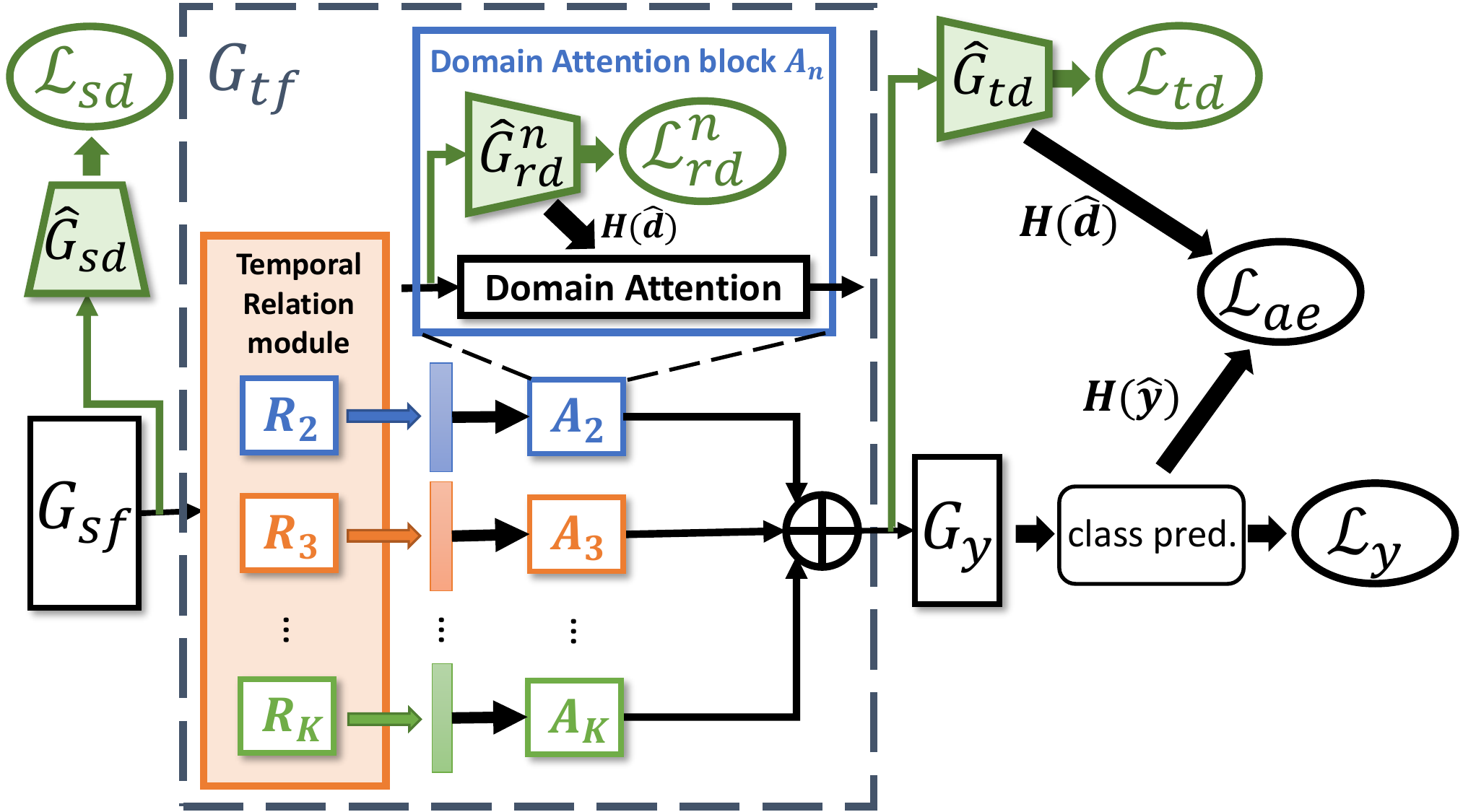}
\caption{The overall architecture of the proposed TA$^3$N. In the temporal relation module, time-ordered frames are used to generate $K$-1 $n$-frame relation feature representations $\textit{\textbf{R}}=\{R_2, ..., R_K\}$.
The attentive entropy loss $\mathcal{L}_{ae}$, which is calculated by domain entropy $H(\hat{d})$ and class entropy $H(\hat{y})$, aims to enhance the certainty of those videos that are more similar across domains. 
}
\label{fig:TemRelation_RevGrad_TransAttn}
\end{figure}

To address this, we propose \textbf{Temporal Attentive Adversarial Adaptation Network (TA$^3$N)}, as shown in \Cref{fig:TemRelation_RevGrad_TransAttn}, by introducing the \textit{domain attention} mechanism, which assigns higher domain attention values to $n$-frame relation features with lower domain entropy, which correspond to higher domains discrepancy. Morevoer, we minimize the class entropy for the videos with low domain discrepancy to refine the classifier adaptation.

The overall loss of TA$^3$N can be expressed as follows:
\begin{equation} \label{eq:loss-all}
\small
\begin{split}
&\mathcal{L} = \frac{1}{N_S}\sum_{i=1}^{N_S}\mathcal{L}^i_y + \frac{1}{N_{S\cup T}}\sum_{i=1}^{N_{S\cup T}}\gamma\mathcal{L}^i_{ae} \\
&-\frac{1}{N_{S\cup T}}\sum_{i=1}^{N_{S\cup T}}(\lambda^s\mathcal{L}^{i}_{sd}+\lambda^r\mathcal{L}^{i}_{rd}+\lambda^t\mathcal{L}^{i}_{td})
\end{split}
\end{equation}
where $N_S$ is the total number of source data, $N_{S\cup T}$ equals the number of all data. $i$ represents the $i$-th video. $\lambda^s$, $\lambda^r$ and $\lambda^t$ is the trade-off weighting for each domain loss $\mathcal{L}_{sd}$, $\mathcal{L}_{rd}$ and $\mathcal{L}_{td}$. $\gamma$ is the weighting for the attentive entropy loss $\mathcal{L}_{ae}$. All the weightings are chosen via grid search. 

Our proposed TA$^3$N and TADA~\cite{wang2019transferable} both utilize entropy functions for attention but with different perspectives. TADA aims to focus on the foreground objects for image DA, while TA$^3$N aims to find important and discriminative parts of temporal dynamics to align for video DA.

\section{Experiments}
We evaluate on three datasets: UCF-Olympic, UCF-HMDB$_{small}$ and UCF-HMDB$_{full}$, as shown in \Cref{table:dataset}. The adaptation setting is noted as ``Source $\rightarrow$ Target". 

\begin{table}[!t]
\centering
\scriptsize
    \begin{tabular}{c|c|c|c}
     & UCF-HMDB$_{small}$ & UCF-Olympic & UCF-HMDB$_{full}$ \\ \hline
    length (sec.) & 1 - 21 & 1 - 39 & 1 - 33 \\ \hline
    resolution & \multicolumn{3}{c}{UCF: $320\times240$ / Olympic: vary / HMDB: vary$\times240$} \\ \hline
    class \# & 5 & 6 & 12 \\ \hline
    video \# & 1171 & 1145 & 3209 \\ \hline
    \end{tabular}
\caption{The comparison of the video DA datasets.}
\label{table:dataset}
\end{table}

\subsection{Datasets and Setup}
\noindent\textbf{UCF-Olympic} and \textbf{UCF-HMDB$\boldsymbol{_{small}}$}:
First, we evaluate our approaches on two publicly used datasets, UCF-Olympic and UCF-HMDB$_{small}$, and compare with all other works that also evaluate on these two datasets~\cite{sultani2014human, xu2016dual, jamal2018deep}. 

\noindent\textbf{UCF-HMDB$\boldsymbol{_{full}}$}:
To ensure large domain discrepancy to investigate DA approaches, 
we build \textbf{UCF-HMDB$\boldsymbol{_{full}}$}, which is around 3 times larger than UCF-HMDB$_{small}$ and UCF-Olympic, as shown in \Cref{table:dataset}. 
To compare with image-based DA approaches, we extend several state-of-the-art methods~~\cite{ganin2015unsupervised, long2017deep, li2018adaptive, saito2018maximum} for video DA with our TemPooling and TemRelation architectures, as shown in \Cref{table:sota_ucf-hmdb_full}. The difference between the ``Target only" and ``Source only" settings is the domain used for training. 

\subsection{Experimental Results and Discussion} \label{sec:experimental_results}
\noindent\textbf{UCF-Olympic} and \textbf{UCF-HMDB$\boldsymbol{_{small}}$}:
In these two datasets, our approach outperforms all the previous methods,
as shown in Table~\ref{table:sota_ucf-olympic_ucf-hmdb-small} (e.g. at least 9\% absolute difference on ``U $\rightarrow$ H").
These results also show that the performance on these datasets is saturated. With a strong CNN as the backbone architecture, even our baseline architecture (TemPooling) can achieve high accuracy without any DA method.
This suggests that these two datasets are not enough to evaluate more sophisticated DA approaches, so larger-scale datasets for video DA are needed.

\begin{table}[!t]
\centering
\footnotesize
    \begin{tabular}{c|c|c|c|c}
    Source $\rightarrow$ Target & U $\rightarrow$ O & O $\rightarrow$ U & U $\rightarrow$ H & H $\rightarrow$ U \\ \hline
    W. Sultani et al.~\cite{sultani2014human} & 33.33 & 47.91 & 68.70 & 68.67 \\ 
    T. Xu et al. ~\cite{xu2016dual} & 87.00 & 75.00 & 82.00 & 82.00 \\ 
    AMLS (GFK)~\cite{jamal2018deep} & 84.65 & 86.44 & 89.53 & 95.36 \\ 
    AMLS (SA)~\cite{jamal2018deep} & 83.92 & 86.07 & 90.25 & 94.40 \\ 
    DAAA~\cite{jamal2018deep} & 91.60 & 89.96 & - & - \\ \hline
    TemPooling & 96.30 & 87.08 & 98.67 & 97.35 \\ 
    TemPooling + RevGrad~\cite{ganin2015unsupervised} & \textbf{98.15} & 90.00 & \textbf{99.33} & 98.41 \\ \hline\hline 
    Ours (TA$^2$N) & \textbf{98.15} & 91.67 & \textbf{99.33} & \textbf{99.47} \\ 
    Ours (TA$^3$N) & \textbf{98.15} & \textbf{92.92} & \textbf{99.33} & \textbf{99.47} \\ \hline
    \end{tabular}
\caption{The accuracy (\%) comparison on UCF-Olympic and UCF-HMDB$_{small}$ (U: UCF, O: Olympic, H: HMDB). 
} 
\label{table:sota_ucf-olympic_ucf-hmdb-small}
\end{table}

\noindent\textbf{UCF-HMDB$\boldsymbol{_{full}}$}:
It is worth noting that the ``Source only" accuracy on UCF-HMDB$_{full}$ is much lower than UCF-HMDB$_{small}$, 
which implies much larger domain discrepancy in UCF-HMDB$_{full}$.
We now answer the question in Section~\ref{sec:TAAN}:
\textit{Does the video DA problem benefit from encoding temporal dynamics into features?} (details in \Cref{table:sota_ucf-hmdb_full})

For the same DA method, TemRelation outperforms TemPooling in most cases, especially for the gain value (e.g. on ``U $\rightarrow$ H", ``TemRelation+RevGrad" reaches 2.77\% absolute accuracy gain while ``TemPooling+RevGrad" only reaches 0.83\% gain).
By simultaneously aligning and encoding temporal dynamics, TA$^2$N 
outperforms other approaches which are extended from more sophisticated image-based DA methods. 
Finally, with the domain attention mechanism, our proposed \textbf{TA$\boldsymbol{^3}$N} reaches 
78.33\%  (6.66\% gain) on ``U $\rightarrow$ H" and 81.79\% (7.88\% gain) on ``H $\rightarrow$ U", achieving state-of-the-art performance in terms of accuracy and gain, as shown in \Cref{table:sota_ucf-hmdb_full}.

\begin{table}[!t]
\centering
\scriptsize
    \begin{tabular}{c|c|c|c|c}
    S $\rightarrow$ T & \multicolumn{2}{c|}{UCF $\rightarrow$ HMDB} & \multicolumn{2}{c}{HMDB $\rightarrow$ UCF} \\ \hline
    Temporal & \multirow{2}{*}{TemPooling} & \multirow{2}{*}{TemRelation} & \multirow{2}{*}{TemPooling} & \multirow{2}{*}{TemRelation} \\ 
    Module &  &  &  &  \\ \hline
    Target only & 80.56 (-) & 82.78 (-) & 92.12 (-) & 94.92 (-) \\ \hline
    Source only & 70.28 (-) & 71.67 (-) & 74.96 (-) & 73.91 (-) \\ 
    RevGrad~\cite{ganin2015unsupervised} & 71.11 (0.83) & 74.44 (2.77) & 75.13 (0.17) & 74.44 (1.05) \\ 
    JAN~\cite{long2017deep} & 71.39 (1.11) & 74.72 (3.05) & 80.04 (5.08) & 79.69 (5.79) \\ 
    AdaBN~\cite{li2018adaptive} & 75.56 (5.28) & 72.22 (0.55) & 76.36 (1.40) & 77.41 (3.51) \\ 
    MCD~\cite{saito2018maximum} & 71.67 (1.39) & 73.89 (2.22) & 76.18 (1.23) & 79.34 (5.44) \\ \hline
    Ours (TA$^2$N) & N/A & 77.22 (5.55) & N/A & 80.56 (6.66) \\ 
    Ours (TA$^3$N) & N/A & \textbf{78.33} (\textbf{6.66}) & N/A & \textbf{81.79} (\textbf{7.88}) \\ \hline
    \end{tabular}
\caption{The comparison of accuracy (\%) for other approaches on UCF-HMDB$_{full}$. The gain values are in ().  
}
\label{table:sota_ucf-hmdb_full}
\end{table}

\noindent\textbf{Integration of $\boldsymbol{\hat{G}_d}$}.
Here we investigate how temporal modules affect each $\hat{G}_d$ without considering the attention mechanism.
For the TemRelation architecture, $\hat{G}_{td}$ integration outperforms $\hat{G}_{sd}$ integration (averagely 0.58\% absolute gain improvement across two tasks), while the TemPooling does not show improvement, as shown in \Cref{table:experiments_dann_position}. This implies that TemRelation better encodes temporal dynamics than TemPooling. 
Finally, by combining all $\hat{G}_d$, the performance improves even more (4.20\% improvement).

\begin{table}[!t]
\centering
\scriptsize
    \begin{tabular}{c|c|c|c|c}
    S $\rightarrow$ T & \multicolumn{2}{c|}{UCF $\rightarrow$ HMDB} & \multicolumn{2}{c}{HMDB $\rightarrow$ UCF} \\ \hline
    Temporal & \multirow{2}{*}{TemPooling} & \multirow{2}{*}{TemRelation} & \multirow{2}{*}{TemPooling} & \multirow{2}{*}{TemRelation} \\ 
    Module &  &  &  &  \\ \hline
    $\hat{G}_{sd}$ & 71.11 (0.83) & 74.44 (2.77) & 75.13 (0.17) & 74.44 (1.05) \\
    $\hat{G}_{td}$ & 71.11 (0.83) & 74.72 (3.05) & 75.13 (0.17) & 75.83 (1.93)  \\ \hline
    All $\hat{G}_d$ & 71.11 (0.83) & \textbf{77.22} (\textbf{5.55}) & 75.13 (0.17) & \textbf{80.56} (\textbf{6.66}) \\ \hline
    \end{tabular}
\caption{The full evaluation of accuracy (\%) for integrating $\hat{G}_d$ in different positions without the attention mechanism. 
}
\label{table:experiments_dann_position}
\end{table}

\begin{figure}[!t]
  \begin{subfigure}[b]{0.235\textwidth}
    \includegraphics[width=\textwidth]{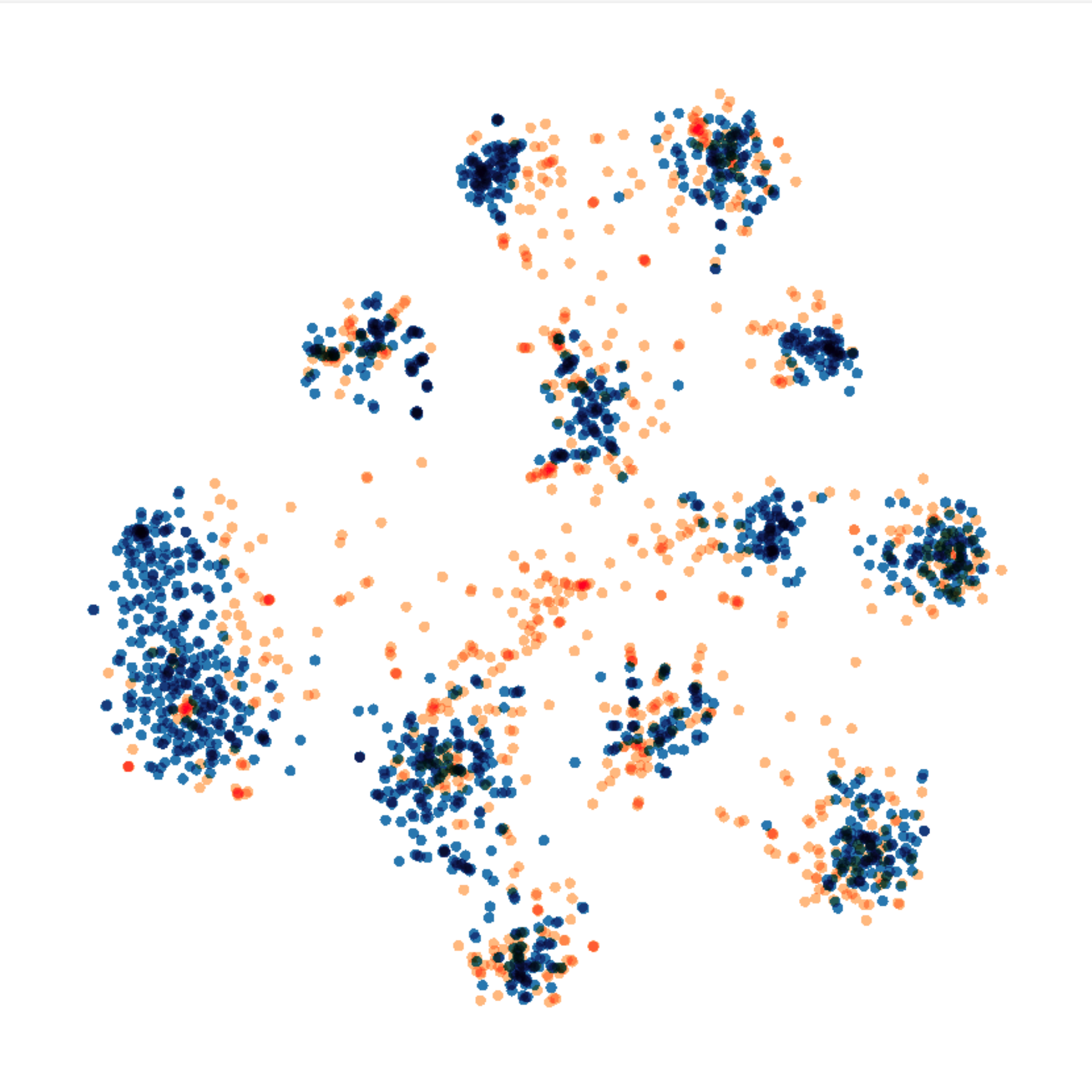}
    \caption{TemPooling + RevGrad~\cite{ganin2015unsupervised}}
    \label{fig:tSNE_TemPooling_G}
  \end{subfigure}
  \begin{subfigure}[b]{0.235\textwidth}
    \includegraphics[width=\textwidth]{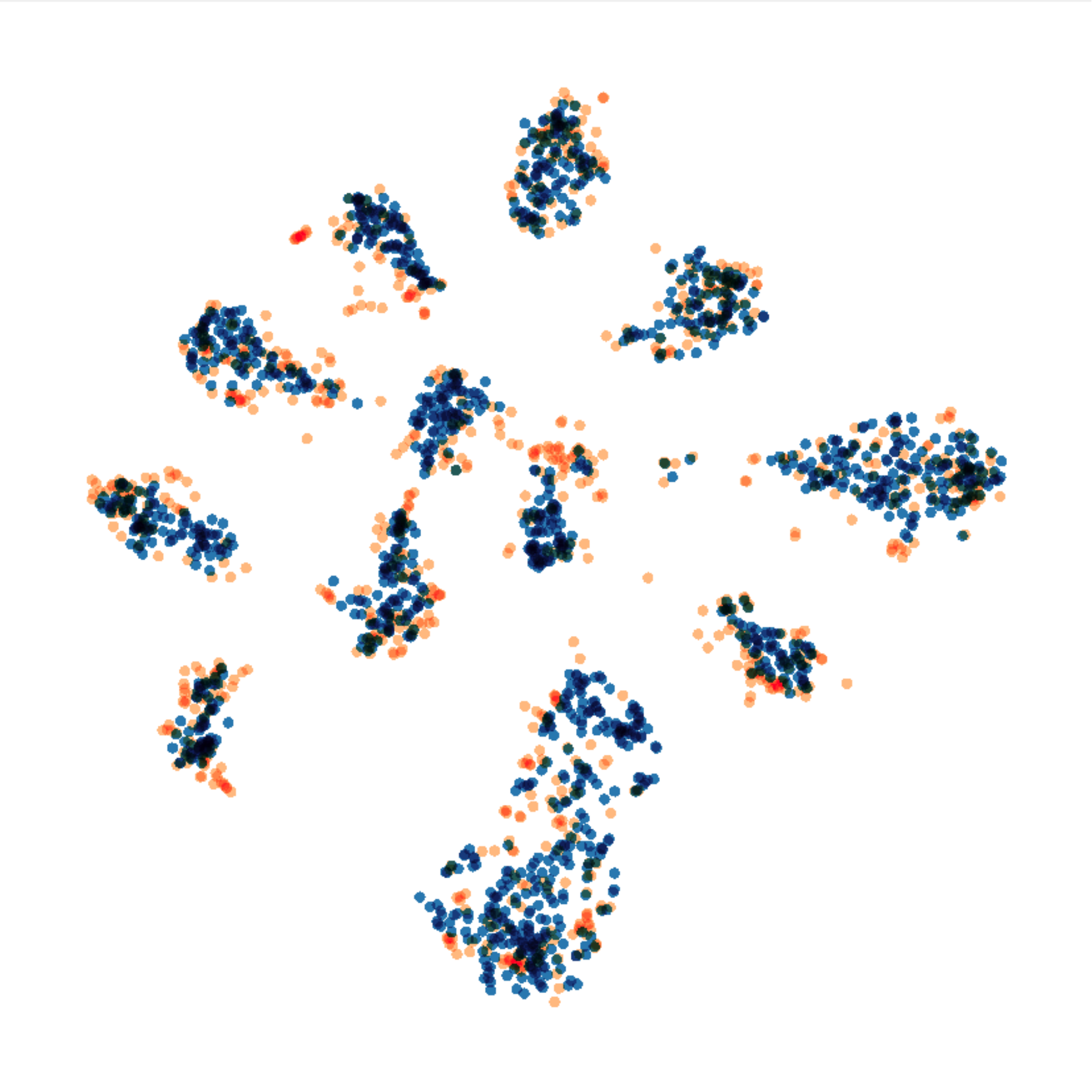}
    \caption{TA$^3$N}
    \label{fig:tSNE_TAAAN}
  \end{subfigure}
\caption{The comparison of t-SNE visualization. The blue and orange dots represent source and target data. 
}
\label{fig:tSNE}
\end{figure}

\noindent\textbf{Visualization of distribution}.
Figure~\ref{fig:tSNE} shows that TA$^3$N 
can group source data (blue dots) into denser clusters and generalize the distribution into the target domains (orange dots) as well, outperforming the baseline model. 

{\small
\bibliographystyle{ieee_fullname}
\bibliography{egbib}

\begin{thebibliography}{10}\itemsep=-1pt

\bibitem{csurka2017comprehensive}
Gabriela Csurka.
\newblock A comprehensive survey on domain adaptation for visual applications.
\newblock In {\em Domain Adaptation in Computer Vision Applications}, pages
  1--35. Springer, 2017.

\bibitem{ganin2015unsupervised}
Yaroslav Ganin and Victor Lempitsky.
\newblock Unsupervised domain adaptation by backpropagation.
\newblock In {\em International Conference on Machine Learning (ICML)}, 2015.

\bibitem{he2016deep}
Kaiming He, Xiangyu Zhang, Shaoqing Ren, and Jian Sun.
\newblock Deep residual learning for image recognition.
\newblock In {\em IEEE conference on Computer Vision and Pattern Recognition
  (CVPR)}, 2016.

\bibitem{jamal2018deep}
Arshad Jamal, Vinay~P Namboodiri, Dipti Deodhare, and KS Venkatesh.
\newblock Deep domain adaptation in action space.
\newblock In {\em British Machine Vision Conference (BMVC)}, 2018.

\bibitem{karpathy2014large}
Andrej Karpathy, George Toderici, Sanketh Shetty, Thomas Leung, Rahul
  Sukthankar, and Li Fei-Fei.
\newblock Large-scale video classification with convolutional neural networks.
\newblock In {\em IEEE conference on Computer Vision and Pattern Recognition
  (CVPR)}, 2014.

\bibitem{kuehne2011hmdb}
Hildegard Kuehne, Hueihan Jhuang, Est{\'\i}baliz Garrote, Tomaso Poggio, and
  Thomas Serre.
\newblock Hmdb: a large video database for human motion recognition.
\newblock In {\em IEEE International Conference on Computer Vision (ICCV)},
  2011.

\bibitem{li2018adaptive}
Yanghao Li, Naiyan Wang, Jianping Shi, Xiaodi Hou, and Jiaying Liu.
\newblock Adaptive batch normalization for practical domain adaptation.
\newblock {\em Pattern Recognition}, 80:109--117, 2018.

\bibitem{long2017deep}
Mingsheng Long, Han Zhu, Jianmin Wang, and Michael~I Jordan.
\newblock Deep transfer learning with joint adaptation networks.
\newblock In {\em International Conference on Machine Learning (ICML)}, 2017.

\bibitem{quionero2009dataset}
Joaquin Quionero-Candela, Masashi Sugiyama, Anton Schwaighofer, and Neil~D
  Lawrence.
\newblock {\em Dataset Shift in Machine Learning}.
\newblock The MIT Press, 2009.

\bibitem{saito2018maximum}
Kuniaki Saito, Kohei Watanabe, Yoshitaka Ushiku, and Tatsuya Harada.
\newblock Maximum classifier discrepancy for unsupervised domain adaptation.
\newblock In {\em IEEE conference on Computer Vision and Pattern Recognition
  (CVPR)}, 2018.

\bibitem{soomro2012ucf101}
Khurram Soomro, Amir~Roshan Zamir, and Mubarak Shah.
\newblock Ucf101: A dataset of 101 human actions classes from videos in the
  wild.
\newblock {\em arXiv preprint arXiv:1212.0402}, 2012.

\bibitem{sultani2014human}
Waqas Sultani and Imran Saleemi.
\newblock Human action recognition across datasets by foreground-weighted
  histogram decomposition.
\newblock In {\em IEEE conference on Computer Vision and Pattern Recognition
  (CVPR)}, 2014.

\bibitem{wang2019transferable}
Ximei Wang, Liang Li, Weirui Ye, Mingsheng Long, and Jianmin Wang.
\newblock Transferable attention for domain adaptation.
\newblock In {\em AAAI Conference on Artificial Intelligence (AAAI)}, 2019.

\bibitem{xu2016dual}
Tiantian Xu, Fan Zhu, Edward~K Wong, and Yi Fang.
\newblock Dual many-to-one-encoder-based transfer learning for cross-dataset
  human action recognition.
\newblock {\em Image and Vision Computing}, 55:127--137, 2016.

\bibitem{zhou2018temporalrelation}
Bolei Zhou, Alex Andonian, Aude Oliva, and Antonio Torralba.
\newblock Temporal relational reasoning in videos.
\newblock In {\em European Conference on Computer Vision (ECCV)}, 2018.

\end{thebibliography}
}

\end{document}